\documentclass{article}

\PassOptionsToPackage{numbers, compress}{natbib}

\usepackage[preprint]{neurips_2019}

\usepackage[utf8]{inputenc} 
\usepackage[T1]{fontenc}    
\usepackage{hyperref}       
\usepackage{url}            
\usepackage{booktabs}       
\usepackage{amsfonts}       
\usepackage{nicefrac}       
\usepackage{microtype}      

\usepackage{algorithmic} 
\usepackage[linesnumbered,ruled,commentsnumbered]{algorithm2e}
\usepackage{bm}
\usepackage{amsmath}
\usepackage{graphicx}
\usepackage{wrapfig}

\usepackage{multirow}

\usepackage{amssymb}
\usepackage{diagbox}
\usepackage[usenames]{color}
\usepackage{colortbl,booktabs}
\usepackage{enumerate}

\title{Relational Reasoning using Prior Knowledge for Visual Captioning}

\author{Jingyi Hou\textsuperscript{1}, Xinxiao Wu\textsuperscript{1}, Yayun Qi\textsuperscript{1}, Wentian Zhao\textsuperscript{1}, Jiebo Luo\textsuperscript{2}, and Yunde Jia\textsuperscript{1}\\
1. Beijing Laboratory of Intelligent Information Technology, School of Computer Science,\\ Beijing Institute of Technology, Beijing 100081, China\\
2. Department of Computer Science, University of Rochester, Rochester NY 14627, USA }

\begin{document}

\maketitle

\begin{abstract}
Exploiting relationships among objects has achieved remarkable progress in interpreting images or videos by natural language. Most existing methods resort to first detecting objects and their relationships, and then generating textual descriptions, which heavily depends on pre-trained detectors and leads to performance drop when facing problems of heavy occlusion, tiny-size objects and long-tail in object detection. In addition, the separate procedure of detecting and captioning results in semantic inconsistency between the pre-defined object/relation categories and the target lexical words. We exploit prior human commonsense knowledge for reasoning relationships between objects without any pre-trained detectors and reaching semantic coherency within one image or video in captioning. The prior knowledge (e.g., in the form of knowledge graph) provides commonsense semantic correlation and constraint between objects that are not explicit in the image and video, serving as useful guidance to build semantic graph for sentence generation. Particularly, we present a joint reasoning method that incorporates 1) commonsense reasoning for embedding image or video regions into semantic space to build semantic graph and 2) relational reasoning for encoding semantic graph to generate sentences. Extensive experiments on the MS-COCO image captioning benchmark and the MSVD video captioning benchmark validate the superiority of our method on leveraging prior commonsense knowledge to enhance relational reasoning for visual captioning.
\end{abstract}

\vspace{-0.2cm}
\section{Introduction}
\label{intro}
\vspace{-0.2cm}
Most of the prominent methods for visual captioning \cite{DBLP:conf/cvpr/DonahueHGRVDS15,DBLP:conf/naacl/VenugopalanXDRM15,DBLP:conf/iccv/VenugopalanRDMD15,DBLP:conf/cvpr/PanXYWZ16} are based on the encoder-decoder framework which directly translates the visual features into sentences, without exploiting high-level semantic entities (e.g., objects, attributes and concepts) as well as relations (e.g., correlation, constraint) among them. 
Recent work \cite{yao2018exploring,li2019know,yang2019cvpr} has shown promising efforts using scene graphs that provide deeper understanding of semantic relationships in images for captioning. These methods usually use pre-trained object detectors to extract scene graph and then reason about object relationships in the graph. This paradigm heavily depends on diverse detectors and leads to substantial performance drop when facing detection challenges such as heavy occlusion, tiny-size objects and long-tail problem. However, human beings can still describe image or video content by summarizing object relationships when some objects are not precisely identified or even absent because of the remarkable reasoning ability resorting to commonsense knowledge. This inspires us to explore how to leverage external prior knowledge for relational reasoning in visual captioning, which mimics human reasoning procedure. 

\begin{figure}[t]
\centering
\includegraphics[width=0.85\textwidth]{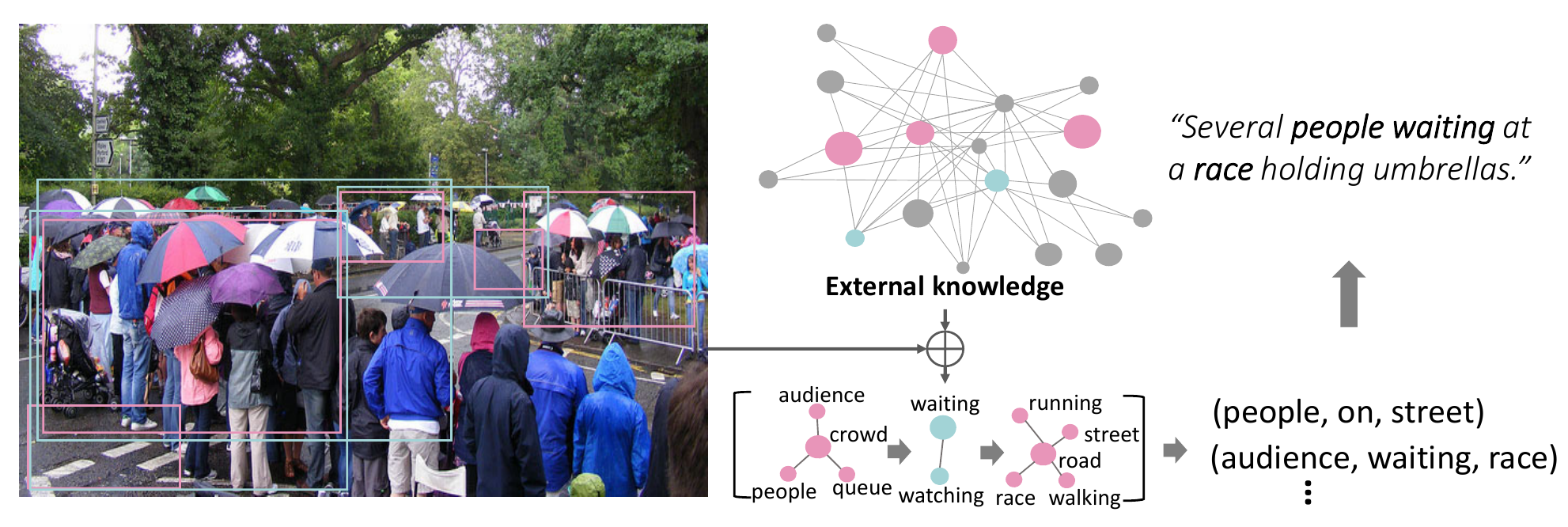}
\vspace{-0.1cm}
\caption{An example of how commonsense reasoning facilitates visual captioning in our work. The concept ``race'' that is absent in the image can be inferred from the visual information and external knowledge via commonsense reasoning.}
\label{fig:fig1}
\end{figure}

As an augmentation of the object relationships explicitly inferred from image or video, the commonsense knowledge about object relationships in the world provides information that is not available in the image or video. 
For example, as shown in Figure~\ref{fig:fig1}, the caption of ``Several people waiting at a race holding umbrellas'' will be generated via prior knowledge when describing a crowd of people standing along the road, even if the image shows no players or running actions (perhaps because the game is yet to begin). Obviously, the relationship of ``people waiting race'' is inferred from the commonsense relationship between ``people'' and ``race'' rather than from the image.
Therefore, it is beneficial to incorporate prior commonsense knowledge with visual information to reason relationships for generating accurate and reasonable captions. 

In this paper, we propose a novel approach of leveraging external prior knowledge to guide reasoning object relationships for image and video captioning.
To augment visual information extracted from images or videos, the prior knowledge provides commonsense semantic correlations and constraints between objects, which are summarized with linguistic knowledge.
An external knowledge graph, i.e., Visual Genome \cite{DBLP:journals/ijcv/KrishnaZGJHKCKL17}, is thus employed as the form of prior knowledge, where the nodes represent the objects and the edges denote the relations between nodes. 
To effectively apply the prior knowledge into visual captioning, we propose a joint reasoning method that incorporates both commonsense reasoning and relational reasoning, and implements them simultaneously. 
The commonsense reasoning aims to select and map local image or video regions into high-level semantic space to build semantic graph via the semantic constraints about relations in the knowledge graph. The relational reasoning is responsible for encoding the semantic graph by refining the representations of regions through a Graphic Convolutional Network (GCN) to generate textual description. To be specific, an iterative reasoning algorithm is developed to alternate between semantic graph generation via commonsense reasoning and visual captioning via relational reasoning.

Our joint reasoning method does not rely on any pre-trained detectors, and does not require any annotations of semantic graph for training. The difficult or even absent objects and relationships that can hardly learned from visual cues can be identified by discovering the inherent relationships guided by external knowledge. 
Another merit of our method lies in the ability of reaching semantic coherency within one image or video for captioning so that the problem of semantic inconsistency between the pre-defined object/relation categories and the target lexical words in existing methods \cite{yao2018exploring,li2019know,yang2019cvpr,DBLP:journals/corr/AdityaYBFA15,DBLP:conf/wacv/ZhouSH19} can be alleviated.  

\vspace{-0.2cm}
\section{Related work}
\label{RW}
\vspace{-0.2cm}
Exploiting relationships between objects for image captioning has gain increasing attentions in nearly a year.
Yao et al. \cite{yao2018exploring} employ two Graphic Convolutional Networks (GCN) to reason the semantic and spatial correlations among visual features of the detected objects and relationships, and add them to a language model to boost image captioning.
\cite{li2019know} generates scene graphs of images by detectors, and builds a hierarchical attention-based model to reason visual relationships for image captioning.
Yang et al. \cite{yang2019cvpr} incorporate language inductive bias into a GCN based image captioning model to not only reason relationship via the GCN but also represent visual information in language domain via a scene graph auto-encoder for easier translation.
The above methods explicitly exploit high-level semantic concepts for image captioning with the pre-defined scene graph of each image and the annotations of object and relationship locations in the image.
Different from them, our method leverages prior knowledge to generate graphs of latent semantic concepts in images  or videos without any pre-trained detectors. 
This enables scene graph generation and visual captioning to be trained in an end-to-end manner, and alleviates the semantic inconsistency in vision-to-language translation.

Some recent methods apply external knowledge graph for image captioning.
In \cite{DBLP:journals/corr/AdityaYBFA15}, commonsense reasoning is used to detect the scene description graph (SDG) of the image, and the SDG can be directly translated into a sentence via a template-based language model.
CNet-NIC \cite{DBLP:conf/wacv/ZhouSH19} incorporates knowledge graphs to augment information extracted from images for captioning.
Different from these methods that directly adds explicit high-level semantic concepts from external knowledge, our method use external knowledge to reason relationships between semantic concepts via joint commonsense and relation reasoning, without facing the  ``hallucinating'' problem as stated in \cite{DBLP:conf/emnlp/RohrbachHBDS18}.

Some Visual Question Answering (VQA) methods \cite{DBLP:conf/emnlp/BerantCFL13,DBLP:conf/kdd/FaderZE14,DBLP:conf/cvpr/SuZDCCL18,DBLP:journals/corr/abs-1904-12584} apply commonsense or relational reasoning.
However, conducting reasoning for visual captioning is more challenging than for VQA. The reason is that for visual captioning the semantic graph is extracted  only from the input  visual cues, while for VQA almost the entire semantic graph is given in terms of the question sentences. 
In this paper, we resort to prior knowledge to tackle the  reasoning problem in visual captioning via a newly proposed joint reasoning method. 
\vspace{-0.2cm}
\section{Method}
\label{PA}
\vspace{-0.2cm}
\subsection{Overview}
\begin{figure}[t]
\centering
\includegraphics[width=0.9\textwidth]{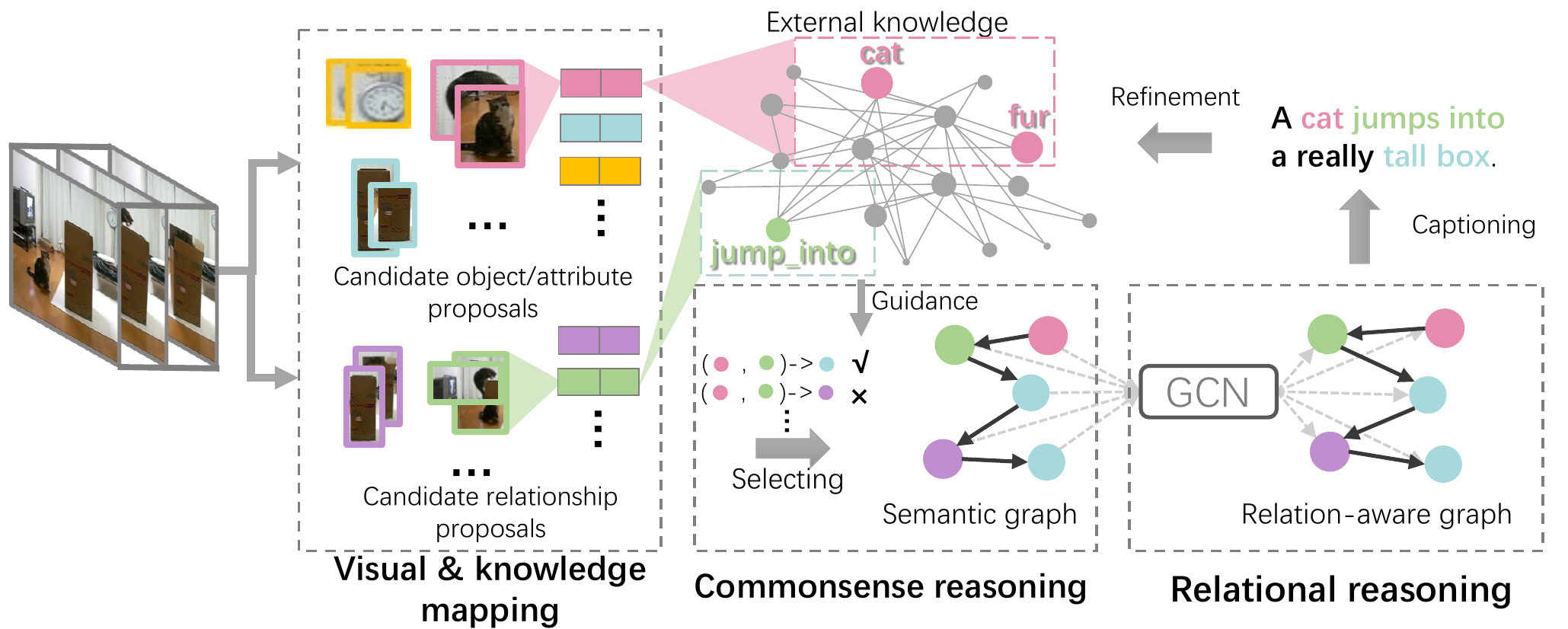}
\vspace{-0.1cm}
\caption{Overview of our method. Our method first assigns regions densely sampled from the input and projects them into low dimensional vectors via visual and knowledge mapping. Then a semantic graph is derived from the learned features with the guidance of the external knowledge graph (knowledge graph). The semantic graph is further represented via relational reasoning for visual captioning, and the captioning result refines the knowledge graph in turn.}
\label{fig:framework}
\end{figure}

\vspace{-0.2cm}
As shown in Figure~\ref{fig:framework}, our method contains three modules: visual and knowledge mapping, commonsense reasoning, and relational reasoning.
In the visual and knowledge mapping module, we first densely sample local image or video regions and these regions are clustered to generate candidate proposals. Then we use visual and knowledge mapping to learn visual feature vectors and knowledge vectors for the candidate proposals, respectively. In the commonsense reasoning module, given the input candidate proposals, the semantic graph is built under the guidance of the external knowledge graph. In the relational reasoning module, given the semantic graph,  the textual description is generated via the GCN and sequence-based language model. In the following, we illustrate our method for the video captioning task as an example for clarity. 

\subsection{Visual and knowledge mapping}
\paragraph{Visual mapping.}
Since visual appearances are critical information of candidate proposals, we extract vectors of visual features via the visual mapping module.
The regions of videos are first densely sampled to fully discover various semantic information from the input video.
To precisely describe the video, three kinds of semantic information are expected to be represented, i.e., objects, attributes, and relationships.
The object and attribute information are represented by using the sampled regions, and the relationship information is represented by using the union areas of two sampled regions.
The visual mapping is then implemented by clustering features extracted from the sampled regions via pre-trained CNNs and using the cluster centers as visual feature vectors of the candidate proposals.
We denote the visual feature vectors as $\bm{V} = [\bm{v}_1,\dots,\bm{v}_{N_v}] \in \mathbb{R}^{L_{v} \times N_v}$ 
where $\bm{v}_i$ represents the $L_{v}$-dimensional visual feature vector of the $i$-th candidate proposal and $N_{v}$ represents the number of candidate proposals in a video. 

\paragraph{Knowledge mapping.} 
The knowledge mapping procedure takes visual feature vectors $\bm{V}$ and knowledge embedding vectors of external knowledge as input, and outputs knowledge vectors of candidate proposals.
The knowledge embedding vectors are calculated based on a knowledge graph on the Visual Genome \cite{DBLP:journals/ijcv/KrishnaZGJHKCKL17} via complEX \cite{DBLP:conf/icml/TrouillonWRGB16}.
Note that the Visual Genome is a large scale dataset containing images annotated by triples of semantic concepts (i.e., objects, attributes, and relationships), but we construct the knowledge graph by only using the triples and not using the images and their bounding box annotations.
Supposing that there are totally $C$  semantic concepts in the knowledge graph, the $L_{k}$-dimensional knowledge embedding vectors are represented as $\bm{E} = [\bm{e}_1,\bm{e}_2,...,\bm{e}_C] \in \mathbb{C}^{L_{k} \times C}$ where $\mathbb{C}$ is the complex domain, which enables the knowledge embedding vectors to represent a directed knowledge graph.
The knowledge vectors $\bm{K} = [\bm{k}_1,\dots,\bm{k}_{N_v}] \in \mathbb{C}^{L_{k} \times N_v}$ of the candidate proposals can thus be derived from the aggregation of the knowledge embedding vectors weighted by the learned soft-assignment.

Specifically, our method learns the weights of the soft-assignment by using multi-label classification models,
and we build three non-linear mapping networks to soft-assign the visual feature vectors with high-level semantic concept labels in the aspects of object, relationship, and attribute in the knowledge graph, respectively.
Since the labels of each visual feature vector are not available, the proposed networks learn labels of the aggregated visual feature vectors by using attention operations during training, and predict the category probabilities of each vector during inference.
The training and inference procedures of the non-linear mapping networks for object, relationship, and attribute are similar, and only the details of training and inference for object is described below.

During training, the ground-truth objects are labeled by the nouns in the captions of the input video. 
The multiple self-attention mechanism is applied to the visual feature vectors to force the network to focus on the relevant vectors to the ground-truth.
Given the visual feature vectors of a video, $K$ attention operations are applied to obtain vectors $\bm{Z} = [\bm{z}_1,\bm{z}_2,...,\bm{z}_K]$.
The $\bm{z}_k$ represents the vector after the $k$-th attention operation given by $\bm{z}_k = \bm{V}\bm{a}_k^\textsf{T}$, where $\bm{a}_k=[a_k^1,\cdots,a_k^V]$ stands for the $k$-th attention weights.
We put $K$ attention weights together to form as $\bm{A} = [\bm{a}_1^\textsf{T}, \bm{a}_2^\textsf{T}, ..., \bm{a}_K^\textsf{T}]^\textsf{T}$ which is  calculated as 
$\bm{A} = \textrm{sparsemax}(\hat{\bm{A}})$, $\hat{\bm{A}} = \bm{W}_2\cdot\textrm{tanh}(\bm{W}_1\bm{V}+\bm{b}_1)$,
where $\bm{W}_1$ and $\bm{W}_2$ are the transformation matrices, $\bm{b}_1$ is the bias, and the $\textrm{sparsemax}(\cdot)$ denotes the sparsemax operation \cite{DBLP:conf/icml/MartinsA16} which is performed along the horizonal dimension of the input matrix.
To determine the object labels of the vectors after attention, the vectors are mapped into $C$ dimensional space by a linear mapping function $f(\cdot)$.
By endowing a sigmoid action function $\sigma(\cdot)$ on the sum of the probabilities of the $K$ attention vectors, we obtain the predicted multi-label semantic probabilities: $\sigma(\sum_{i=1}^{K}f(\bm{z}_{i})) \in \mathbb{R}^{C \times 1}$.
The NLTK toolkit \cite{DBLP:journals/nle/Xue11} is used to tag the parts-of-speech on the words in the ground-truth sentence, which gives the labels of the multiple objects in the video.
With the predicted probabilities and ground-truth labels, the non-linear mapping network for object is trained under the guidance of the binary cross entropy loss function.
In addition, in order to encourage the model to focus on diverse semantic concepts in each video, we set a constraint to regularize the vectors $\bm{Z}$, given by $-\sum_{i\neq j}\textrm{KL}(p(z_i)||p(z_j))$, where $\textrm{KL}(\cdot)$ calculates the Kullback–Leibler divergence, and $p(\cdot)$ is the softmax function.

During inference, the visual feature vector of the $i$-th proposal  is fed into the non-linear mapping network without the attention operation, i.e., $\bm{z}_i = \bm{v}_i$.
We apply the sparsemax operation to normalize $f(\bm{z}_i)$ into the probabilities, $\bm{p}_i=\textrm{sparsemax}(f(\bm{z}_i))$, of object predictions as the weights of embedding vectors of the knowledge graph.
Consequently, the knowledge vector $\bm{k}_{i}^{obj} \in \mathbb{R}^{L_{k} \times 1}$ is calculated as $\bm{k}_{i}^{obj} = \bm{E}\bm{p}_{i}$.
The visual feature vectors and knowledge vectors of objects in a video can thus be obtained as $\mathcal{V}^{obj} = \{\bm{v}_{i}|_{i=1}^{N_{v}}\}$, $\mathcal{K}^{obj} = \{\bm{k}_{i}^{obj}|_{i=1}^{N_{v}}\}.$

The non-linear mapping networks for relationships and attributes are trained in a similar manner except for the ground-truth labels, and the input of the network for the relationship.
The final visual feature vectors and knowledge vectors of relationships and attributes are represented as $\mathcal{V}^{rel} = \{\bm{r}_{i}|_{i=1}^{F_{r}}\}$, $\mathcal{K}^{rel} = \{\bm{k}_{i}^{rel}|_{i=1}^{F_{r}}\}$, $\mathcal{V}^{att} = \{\bm{v}_{i}|_{i=1}^{N_{v}}\}$, and $\mathcal{K}^{att} = \{\bm{k}_{i}^{att}|_{i=1}^{N_{v}}\}$,
where $F_{r}$ is the number of cluster centers on relationship visual feature vectors, and the knowledge vectors $\bm{k}_{i}^{rel}$ and $\bm{k}_{i}^{att}$ are derived from the weighted aggregations of the relationship and attribute embedding vectors, respectively.
The final visual feature vector and knowledge vector sets of candidate proposals of a video are described as $\mathcal{V} = \mathcal{V}^{obj} \cup \mathcal{V}^{rel} \cup \mathcal{V}^{att}$ and $\mathcal{K} = \mathcal{K}^{obj} \cup \mathcal{K}^{rel} \cup \mathcal{K}^{att}$.

\begin{figure}[t]
\centering
\includegraphics[width=0.9\textwidth]{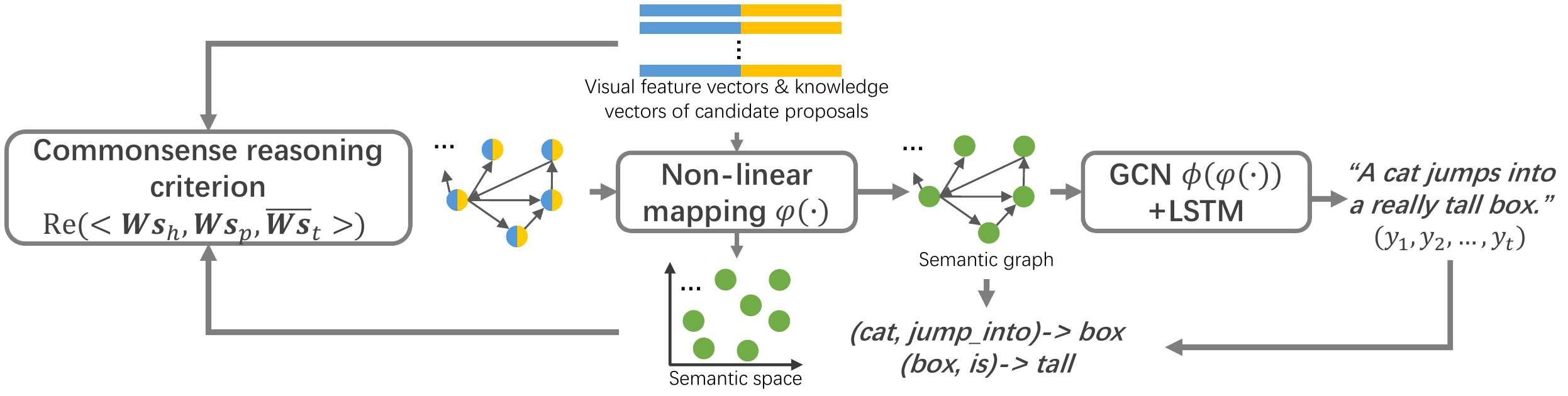}
\vspace{-0.1cm}
\caption{Joint reasoning for visual captioning. The ``non-linear mapping'' block is learned by commonsense reasoning. The ``GCN+LSTM'' block is built for relational reasoning. Both blocks are jointly updated through back propagation. Given the features of candidate proposals, the ``Commonsense reasoning criterion'' block selects semantic features of the graph learned by the ``non-linear mapping'' block.}
\label{fig:jointreason}
\end{figure}

\vspace{-0.2cm}
\subsection{Joint reasoning for visual captioning}
Given the visual feature vectors and knowledge vectors of all the candidate proposals, i.e., $\mathcal{V}\cup\mathcal{K}$, generated by visual and knowledge mapping, we conduct visual captioning by using commonsense reasoning and relational reasoning, as illustrated in Figure~\ref{fig:jointreason}.
The commonsense reasoning maps the concatenation of visual feature vectors and knowledge vectors into a semantic space to obtain the semantic features $\varphi(\bm{v}_i,\bm{k}_i)$, and constructs a semantic graph using selected semantic features with the guidance of triplet constraints summarized in the knowledge graph.
The relational reasoning module learns the relation-aware feature $\phi(\varphi(\bm{v}_i,\bm{k}_i))$ for each vertex via a GCN, and generates a description using a sequence-based language model given the input soft-assigned relation-aware features.

\paragraph{Commonsense reasoning.}
Taking visual feature vectors $\mathcal{V}$ and knowledge vectors $\mathcal{K}$ as input, we further represent the candidate proposals as latent semantic concepts $\mathcal{S}$ by using a non-linear mapping function: $\bm{s}_i = \varphi(\bm{v}_i,\bm{k}_i)$, $\bm{s}_i \in \mathcal{S}$.
The non-linear mapping function is updated with the back-propagation of the entire joint reasoning framework.
The semantic features are learned to satisfy that the correlations and constrains among the object, relationship and attribute vertices can be inferred by a commonsense reasoning criterion to generate the semantic graph of the input video. 
Different from some visual relationship detection methods \cite{DBLP:conf/cvpr/LiangLX17,DBLP:conf/eccv/LuKBL16,DBLP:conf/cvpr/ZhangKCC17} that leverage language prior or regularize relation embedding space to improve the performance, 
our method applies commonsense reasoning in latent semantic concepts without explicit supervision, so as to generate the most relevant scene graph for describing the video.

Concretely, a knowledge graph is a collection of factual triplets, where each triplet represents a relationship between a head entity and a tail entity.
Let $\mathcal{S} = \mathcal{S}^h\cup\mathcal{S}^r\cup\mathcal{S}^t$, where $\mathcal{S}^h$, $\mathcal{S}^r$ and $\mathcal{S}^t$ are the head, relationship and tail entity sets, respectively.
We learn a commonsense reasoning criterion to represent the semantic features via complex vectors, such that not only the symmetric but also the antisymmetric relations among the semantic concepts can be measured.
Following \cite{DBLP:conf/icml/TrouillonWRGB16}, the criterion measures the real part of the composition of the semantic triplet $(\bm{s}^h, \bm{s}^r, \bm{s}^t)$ for representing the correlation in the triplet:
\begin{eqnarray}
\begin{aligned}
\label{eq:criterion}
&\textrm{Re}( <\bm{W}\bm{s}^h, \bm{W}\bm{s}^r, \overline{\bm{W}\bm{s}^t}> ) \\
& = \ <\textrm{Re}(\bm{W}\bm{s}^h),\textrm{Re}(\bm{W}\bm{s}^r),\textrm{Re}(\bm{W}\bm{s}^t)>+<\textrm{Re}(\bm{W}\bm{s}^h),\textrm{Im}(\bm{W}\bm{s}^r),\textrm{Im}(\bm{W}\bm{s}^t)> \\ 
& \ \ \ \ + <\textrm{Im}(\bm{W}\bm{s}^h),\textrm{Re}(\bm{W}\bm{s}^r),\textrm{Im}(\bm{W}\bm{s}^t)>-<\textrm{Im}(\bm{W}\bm{s}^h),\textrm{Im}(\bm{W}\bm{s}^r),\textrm{Re}(\bm{W}\bm{s}^t)>,
\end{aligned}
\end{eqnarray}
where $\bm{s}^h\in\mathcal{S}^h$, $\bm{s}^r\in\mathcal{S}^r$, and $\bm{s}^t\in\mathcal{S}^t$, $\bm{W}\in $ is a weight matrix that converts the semantic features to complex vectors, $\overline{\bm{W}\bm{s}^t}$ is the complex conjugate of ${\bm{W}\bm{s}^t}$, $<\cdot>$ denotes the multi-linear dot product of the vectors in the triplet, and $\textrm{Re}(\cdot)$ and $\textrm{Im}(\cdot)$ denote the real and imaginary parts of a number, respectively.
Note that the form of the triplet is ordered, and attribute vertices could only be the tail entities.

We select triplets with large responses on the criterion from the candidate proposals to generate the semantic graph of the corresponding video.
In analogy to non-maximum suppression (NMS), we suppress triplets that have more than one same vertex or score with the local maxima measured by the criterion, and eliminate triplets whose scores are lower than $-1$.

\paragraph{Relational reasoning.}
In order to process relational reasoning on the semantic graph generated by the commonsense reasoning, a GCN \cite{DBLP:conf/cvpr/JohnsonGF18} is employed to propagate information along edges of the graph.
The GCN contextually encodes features in the semantic graph to generate relation-aware features.
We add a residual connection to each layer of the GCN to facilitate the optimization of the proposed model, which is different from \cite{DBLP:conf/cvpr/JohnsonGF18}.

As for visual captioning, we process triplets in the output relation-aware graph with the attention mechanism and feed them into a sequence-based language model \cite{DBLP:conf/cvpr/00010BT0GZ18}, where each triplet is the represented by the concatenation of the relation-aware features of the head, relationship, and tail entities.
The input of the top-down attention LSTM layer at each time step is the concatenation of the previous hidden state of the language LSTM layer, the mean-pooled frame-level video features, and the embedding vector of the previously generated word.
The hidden state of the attention LSTM at time step $t$ is respectively fused with the triplet features of the semantic graph after GCN to derive the attention weights, $\bm{\alpha}_t=[\alpha_{1,t},\dots,\alpha_{G,t}]$, where $G$ is the number of triplets in the graph.
We can then easily obtain the attended triplet feature with the attention weights.
The input of the language LSTM layer at each time step consists of the attended triplet feature concatenated with the hidden state of the attention LSTM, and the output is the conditional distribution over the words in the dictionary.

\paragraph{Objective.}
Two losses are effectively combined to train the entire visual captioning model. 
One loss is a cross-entropy loss for generating sentences, and it is defined as
\begin{eqnarray}
\begin{aligned}
\label{eq:xeloss}
L_{c}=-\sum_{t=1}^T\log \big( Pr(y_t|y_{1:t-1},\mathcal{I})\big),
\end{aligned}
\end{eqnarray}
where $Pr(y_t|y_{1:t-1}, \mathcal{I})$ denotes the probability that the prediction is the ground-truth word $y_t$ given the previous word sequence $y_{1:t-1}$ and all the features $\mathcal{I}$ including the frame-level features and candidate proposal features (visual feature vectors and knowledge vectors) of the input videos.

The other loss guides the learning of the semantic features of each vertex to capture correlation information with its adjacent vertices, which is measured by the commonsense reasoning criteria when the semantic features are mapped into the complex domain:
\begin{eqnarray}
\begin{aligned}
\label{eq:seloss}
L_{s}=\sum_{g=1}^{G}\sum^{T}_{t=1}(\alpha_{g,t}-\gamma)\log(1+\exp(-\textrm{Re}(<\bm{W}\bm{s}^h_g,\bm{W}\bm{s}^r_g,\overline{\bm{W}\bm{s}^t_g}>)))+\lambda||\bm{W}||_2^2,
\end{aligned}
\end{eqnarray}
where the parameter $\lambda$ represents the importance of the regularization term, and $\gamma$ is a threshold that determines triplets to be punished.
In the experiments, we set $\lambda=0.01$ and $\gamma=0.3$ empirically.

Consequently, the overall loss is formulated as
\begin{eqnarray}
\begin{aligned}
\label{eq:totalloss}
L = L_{c}+\beta L_{s},
\end{aligned}
\end{eqnarray}
where $\beta$ is a hyper-parameter which can be tuned.
Since $L_{s}$ is constrained on the learning of attention weights $\{\bm{\alpha}_t|t=1,\dots,T\}$ guided by $L_{c}$, we set $\beta$ to $0$ during the first few epochs of training, and $0.1$ afterwards.

\IncMargin{1em}
\begin{algorithm}
\SetKwInOut{INIT}{Initialization}
\footnotesize
\caption{Joint reasoning for visual captioning.}
\label{alg:alg1}
\KwIn{visual feature vectors $\mathcal{V}=\cup_{n=1}^N\mathcal{V}_n$ and knowledge vectors $\mathcal{K}=\cup_{n=1}^N\mathcal{K}_n$ of $N$ images or videos.
}
\KwOut{Joint reasoning model.}
\textbf{Initialization}: $\mathcal{H}_{n} = \mathcal{K}_{n}, \ \forall n = 1, \cdots, N$; \\
\Repeat{Convergence}
{
    $\bullet$ \textbf{Semantic Graph Generation:} \\
    \For{$n=1, \cdots, N$}
    {
        $(\mathcal{V}^S_n,\mathcal{V}^S_n) \Leftarrow$ Select object, relationship and attribute vertices from $(\mathcal{V}_{n},\mathcal{K}_{n})$ by using (\ref{eq:criterion}) on $\mathcal{H}_{n}$; \\
    }
    $\mathcal{V}^S \Leftarrow \cup_{n=1}^N\mathcal{V}^S_n, \ \mathcal{K}^S \Leftarrow \cup_{n=1}^N\mathcal{K}^S_n$; \\
    Non-linearly map $\mathcal{V}^S$ and $\mathcal{K}^S$ into semantic space $\varphi(\mathcal{V}^S,\mathcal{K}^S)$;\\
    $\bullet$ \textbf{Visual Captioning:} \\
    Map $\varphi(\mathcal{V}^S,\mathcal{K}^S)$ into $\phi(\varphi(\mathcal{V}^S,\mathcal{K}^S))$ by using relational reasoning based on GCN;\\
    $\bullet$ \textbf{Update:} \\
    Update parameters of $\phi(\cdot)$ and $\varphi(\cdot)$, and the sequence-based language model by minimizing $L$.\\
    $\mathcal{H}_{n} \Leftarrow \varphi(\mathcal{V}_{n},\mathcal{K}_{n}), \ \forall n = 1, \cdots, N$; \\
 }
\end{algorithm}

\paragraph{Alternative algorithm.}
Although the proposed framework can be trained in an end-to-end manner, we find that learning the commonsense reasoning module faces an optimization challenge: the construction of the semantic graph involves hard assignment operations, i.e., selecting triplets.
To address this issue, we develop an iterative learning algorithm to alternate between semantic graph generation via commonsense reasoning and visual captioning via relational reasoning, as summarized in Algorithm~\ref{alg:alg1}.

\vspace{-0.1cm}
\section{Experiments}
\label{Exper}

\vspace{-0.1cm}
\subsection{Datasets}
We use MSCOCO \cite{DBLP:conf/iccv/VenugopalanRDMD15} for image captioning and  MSVD \cite{DBLP:conf/iccv/GuadarramaKMVMDS13} for video captioning.
On MSCOCO, we follow the split standard of \cite{DBLP:journals/pami/KarpathyF17} which takes 113,287 images for training, 5,000 for validation and 5,000 for testing.
On MSVD, we split the videos into three sets following \cite{DBLP:conf/iccv/VenugopalanRDMD15}, consisting of 1,200 training videos, 100 validation videos and 670 testing videos.
The metrics of BLEU-4 (B@4) \cite{DBLP:conf/acl/PapineniRWZ02}, METEOR \cite{DBLP:conf/wmt/DenkowskiL14}, CIDEr \cite{DBLP:conf/cvpr/VedantamZP15}, and SPICE \cite{DBLP:conf/eccv/AndersonFJG16} are used for evaluations by the MSCOCO toolkit \cite{DBLP:journals/corr/ChenFLVGDZ15}.
For all the metrics, higher values indicate better performances.

\vspace{-0.1cm}
\subsection{Implementation details}
\vspace{-0.1cm}
For image captioning, the visual feature vector of each region is calculated after ROI pooling from the corresponding region in the feature map of the res5c layer of ResNet-101 \cite{he2016deep}.
The visual feature of the whole image is the output of the pool5 layer of ResNet-101.
For video captioning, the visual features of the whole video are derived from C3D and InceptionV4.
Similar to the image captioning, the visual feature vector of each region is calculated by concatenating features after average pooling from the corresponding region in the feature map of the last convolutional layers of C3D and InceptionV4.
The visual feature of each video frame is the concatenation of outputs of the pooling layer after the last convolutional layers of C3D and InceptionV4.
To reduce the computational resource, we employ the RPN \cite{DBLP:journals/pami/RenHG017} without NMS to densely sample candidate object regions with scores higher than threshold 0.7.

During the initialization, the k-means clustering is used.
For data augmentation, we repeatedly conduct k-means operations to obtain multiple groups of candidate proposals from each image or video, and the number of clusters are set from 5 to 10.
The number of the sparse attention operations is set to 3 according to the mAP of the multi-label classification by the non-linear networks for knowledge mapping on the validation set.
In the sequence-based language model, we set the number of hidden units in each LSTM as 512, and set the size of the input word embedding as 512.
During training, the convergence criterion is considered as that the CIDEr score on the validation set stops increasing in 10 consecutive epochs.
During inference, the sizes of beam search are set to 3 and 5 to generate the final sentences in image captioning and video captioning, respectively.

\vspace{-0.2cm}
\subsection{Comparison with the state-of-the-art methods}
\vspace{-0.1cm}
Table~\ref{tab:coco} shows the comparison results between our method and several recent methods that are closely related to our method on the MSCOCO dataset.
All the compared methods \cite{DBLP:conf/cvpr/GanGHPTGCD17,DBLP:conf/cvpr/00010BT0GZ18,yao2018exploring,DBLP:conf/wacv/ZhouSH19,li2019know} leverage explicit high-level semantic concepts (e.g., objects, relationships) for image captioning.
Compared with \cite{DBLP:conf/cvpr/GanGHPTGCD17} which uses multi-label classifier for generating explicit semantic concepts without exploiting their relations, our method achieves better results, which validates that joint reasoning can benefit learning semantic relationships for visual captioning. 
Compared with \cite{DBLP:conf/cvpr/00010BT0GZ18,yao2018exploring,DBLP:conf/wacv/ZhouSH19,li2019know} which use pre-trained detectors to explore visual relationships for captioning, our method without annotations still achieves comparable performances, 
demonstrating that exploiting relationships actually benefits from prior knowledge and does not necessarily rely on pre-trained detectors.

Table~\ref{tab:msvd} shows the comparison results on the MSVD dataset. 
Our method performs better than \cite{DBLP:conf/cvpr/GanGHPTGCD17} for video captioning as well, which further proves the advantages of commonsense reasoning and relational reasoning.
Our method substantially outperforms \cite{DBLP:journals/corr/abs-1902-10322} which detects objects from videos using detectors pre-trained on image dataset.
It indicates that using external knowledge to extract the object information in our method is more general than training object detectors, 
when there exists domain gap between the detection dataset and the captioning dataset. 

\begin{table}\scriptsize
\parbox{.45\linewidth}{
\centering
\caption{Comparison results on MSCOCO.}
\label{tab:coco}
\centering
\begin{tabular}{lccccc}
\toprule
Methods & B@4  & METEOR & CIDEr & SPICE\\
\midrule
SCN  \cite{DBLP:conf/cvpr/GanGHPTGCD17}  & 33.0 & 25.7 & 101.2 & - \\
Up-Down \cite{DBLP:conf/cvpr/00010BT0GZ18}   & 36.2 & 27.0 & 113.5  & 20.3 \\
GCN-LSTM \cite{yao2018exploring}   & \textbf{37.1} & \textbf{28.1} & 117.1  & \textbf{21.1} \\
CNet-NIC \cite{DBLP:conf/wacv/ZhouSH19} & 29.9 & 25.6 & 107.2  & - \\
KMSL \cite{li2019know}   & 33.8 & 26.2 & 110.3  & 19.8 \\
\midrule
Ours & 36.7 & \textbf{28.1} & \textbf{117.3} & 20.1  \\
\bottomrule
\end{tabular}
}
\hfill
\parbox{.45\linewidth}{
\caption{Comparison results on MSVD.}
\label{tab:msvd}
\begin{tabular}{lccc}
\toprule
Methods & B@4  & METEOR  & CIDEr \\
\midrule
SCN  \cite{DBLP:conf/cvpr/GanGHPTGCD17}  & 51.1 & 33.5 & 77.7 \\
GRU-EVE \cite{DBLP:journals/corr/abs-1902-10322}  & 47.8 & 35.0 & 78.1 \\
\midrule
Up-Down \cite{DBLP:conf/cvpr/00010BT0GZ18} & 45.8 & 32.5 & 72.2  \\
Ours w/o CR    & 43.7 & 33.1 & 70.5  \\
Ours w/o RR    & 50.8 & 35.4 & 85.4  \\
Ours    & \textbf{53.3} & \textbf{36.0} & \textbf{89.6} \\
\bottomrule
\end{tabular}
}
\end{table}

\vspace{-0.2cm}
\subsection{Ablation Study}
\vspace{-0.2cm}
Ablation studies are conducted for evaluating the importance of each individual component.
The results on the MSVD dataset are reported in Table~\ref{tab:msvd} with the captioning model of [4] as the backbone.

\textbf{The effect of joint reasoning.} We analyze the effect of joint commonsense and relational reasoning by comparing our method with the Up-Down model \cite{DBLP:conf/cvpr/00010BT0GZ18}.
\begin{wrapfigure}{r}{5cm}
\includegraphics[width=5cm]{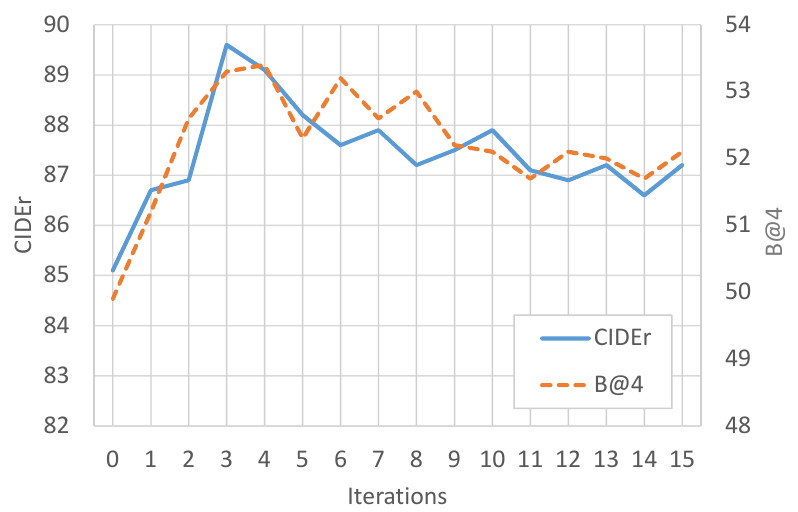}
\caption{Learning curves of the CIDEr and B@4 scores for the analysis of the alternative algorithm.}
\label{fig:curve}
\end{wrapfigure}
For fair comparison, the visual features fed into the Up-Down model are the same with ours.
The model uses Faster R-CNN \cite{DBLP:conf/nips/RenHGS15} to detect spatial regions for generating spatial image features as the input to the bottom-up attention model.
For all the metrics, our method achieves substantially better results than the Up-Down model, which clearly validates that joint reasoning can significantly boost the performance of visual captioning.

\textbf{The effect of commonsense reasoning.}
To analyze the effect of commonsense reasoning, we remove the commonsense reasoning, and instead, apply Faster R-CNN to generate semantic graph, called ``Ours w/o CR''.
The great performance improvement of our method than ``Ours w/o CR'' validates the important effect of commonsense reasoning on finding the most relevant semantic concepts and relationships for captioning.
It is interesting to observe that the performance of ``Ours w/o CR'' is even worse than Up-Down on B@4 and CIDEr, which is mainly because the Faster R-CNN used in the Up-Down model is trained on the MSCOCO dataset which has domain bias with the MSVD dataset.
In contrast, our method does not rely on existing detectors, and the commonsense reasoning in our method is trained in weakly-supervised manner on the MSVD dataset.

\textbf{The effect of relational reasoning.}
We analyze the effect of relational reasoning by removing the GCN from our framework, called ``Ours w/o RR''.
Our method outperforms ``Ours w/o RR'', which shows that the relational reasoning module that encodes the semantic graph is beneficial for boosting the performance of visual captioning. 

\textbf{The effect of alternative algorithm.}
For a more intuitive view of our alternative algorithm, we drew the learning curves of the CIDEr and B@4 scores on the test set of MSVD in Figure~\ref{fig:curve}.
In the first iteration, the model is trained without the loss $L_S$ at the beginning, denoted as iteration 0.
As illustrated in Figure~\ref{fig:curve}, the model converges after three iterations, and the CIDEr score drops afterwards because of the overfitting of the model.


\vspace{-0.3cm}
\section{Conclusion}
\vspace{-0.2cm}
We have presented a joint reasoning method for visual captioning via the external knowledge graph, where commonsense reasoning embedds image or video regions into the semantic space to build a semantic graph and relational reasoning encodes the semantic graph to generate textual descriptions.
Without using pre-defined object and relationship detectors, our method can take full use of the semantic constraints summarized in the knowledge graph to achieve the global semantic coherency within one image or video.
Moreover, our method can be readily extended by integrating multiple chosen knowledge forms for more advanced reasoning which can be easily plugged into captioning networks for endowing its ability in exploiting relationships. 
Experiments on the MS-COCO image captioning and the MSVD video captioning benchmarks demonstrate the superiority of our method for visual captioning.

{\small
\bibliographystyle{plain}
\bibliography{neurips_2019}
}

\end{document}